\documentclass[sigconf,3,nonacm]{acmart}
\usepackage{multirow}
\usepackage{graphicx}
\usepackage{datetime2}
\usepackage{stfloats}
\usepackage{bbding}
\usepackage{fancyhdr}
\fancypagestyle{plain}{
    \fancyhf{}
    \fancyfoot[R]{\thepage}

}
\usepackage{pdfpages}\usepackage{datetime}

\AtBeginDocument{%
  \providecommand\BibTeX{{%
    \normalfont B\kern-0.5em{\scshape i\kern-0.25em b}\kern-0.8em\TeX}}}

\setcopyright{none}
\begin{document}

\fancyfoot{}
\fancyfoot[R]{\thepage}
\renewcommand{\footruleskip}{10pt}

\title{A Survey of Graph and Attention Based Hyperspectral Image Classification Methods for Remote Sensing Data}

% \author{Aryan Vats, Megha Sharma, Reshma Ramachandra, Rynaa Grover}
% \affiliation{%
%   \institution{Georgia Institute of Technology}}
% \email{{avats31,meghasharma,reshmaram,rgrover30}@gatech.edu}

\author{Aryan Vats}
\affiliation{%
  \institution{Georgia Institute of Technology}
\country{}}
\email{avats31@gatech.edu}

\author{Manan Suri}
\affiliation{%
  \institution{Indian Institute of Technology Delhi}
\country{}}
\email{manansuri@ee.iitd.ac.in}
\country{}

\renewcommand{\shortauthors}{Vats, Suri}
\renewcommand{\shorttitle}{A Survey of Graph and Attention Based Hyperspectral Image Classification Methods}
\begin{abstract}
  The use of Deep Learning techniques for classification in Hyperspectral Imaging   (HSI) is rapidly growing and achieving improved performances. Due to the nature of the data captured by sensors that produce HSI images, a  common issue is the dimensionality of the bands that may or may not contribute to the label class distinction. Due to the widespread nature of class labels, Principal Component Analysis is a common method used for reducing the dimensionality. However,there may exist methods that incorporate all bands of the Hyperspectral image with the help of the Attention mechanism. Furthermore, to yield better spectral spatial feature extraction, recent methods have also explored the usage of Graph Convolution Networks and their unique ability to use node features in prediction, which is akin to the pixel spectral makeup. In this survey we present a comprehensive summary of Graph based and Attention based methods to perform Hyperspectral Image Classification for remote sensing and aerial HSI images. We also summarize relevant datasets on which these techniques have been evaluated and benchmark the processing techniques. 
\end{abstract}
\keywords{Remote Sensing, HSI, Graph Convolutional Networks, Attention, U-Nets, Hyperspectral Images}

\affiliation{
  \country{} 
}
\maketitle



% \keywords{Recommendation System, Playlist Continuation, Random Walk, Graph Neural Network, GraphSAGE}


\section{Introduction}
Hyperspectral imaging (HSI) is an advanced remote sensing technique that provides a wealth of spectral information across a wide range of electromagnetic wavelengths, usually greater than 100 bands. The Airborne Visible Infrared Imaging Sensor(AVIRIS) \cite{vane-1993} usually captures 224 bands, used in many of the datasets observed for HSI classification. The Reflective Optics Imaging Spectrometer (ROSIS) \cite{kunkel-1988}, on the other hand captures 115 bands.  It has gained significant attention in various fields, including agriculture, environmental monitoring, mineralogy, and urban planning, due to its ability to capture detailed and fine-grained information about the Earth's surface. HSI sensors can capture hundreds of narrow and contiguous spectral bands, enabling the characterization and discrimination of different materials based on their spectral signatures.

One of the key challenges in HSI analysis is the accurate classification of the acquired hyperspectral data. The classification task involves assigning each pixel in the image to one of several predefined classes or categories. Traditional classification methods, such as support vector machines (SVM) \cite{6521421} \cite{OKWUASHI2020107298} or random forests \cite{CHUNHUI201861}, have been widely used for HSI classification. However, they often struggle to exploit the rich spectral information and the spatial contextual relationships inherent in hyperspectral images.

In recent years, graph-based and attention-based methods have emerged as promising approaches for HSI classification. These methods leverage the inherent spatial and spectral correlations in hyperspectral data to improve classification accuracy. Graph-based methods model the relationships between neighboring pixels as a graph structure and utilize Graph Convolutional Networks \cite{DBLP:journals/corr/KipfW16} (GCNs) or Graph Attention Networks \cite{velickovic-2017} (GATs) to capture the dependencies and propagate information across the graph. On the other hand, attention-based methods focus on learning the relevant spectral bands or spatial locations by assigning attention weights to different inputs, enabling the model to selectively attend to discriminative features.

This survey paper aims to provide a comprehensive overview of the state-of-the-art graph-based and attention-based methods for HSI classification. We will review and analyze significant and recent works in the field, highlighting their key contributions, methodologies, and performance. By surveying the existing literature, this paper intends to serve as a valuable resource for researchers and practitioners interested in exploring and advancing the field of graph and attention-based methods for HSI classification.

Section \ref{section2} explores the various HSI datasets used by the models assessed, and contrasts the dataset features. Section \ref{section3} assesses the Graph Based and Attention Based techniques utilized to perform HSI classification, with focus on each specific model.  Section \ref{section4} benchmarks the results of each model, assessing common ground and making observations on trends.
% In this paper we aim to 

% \begin{itemize}
%     \item Analyse works that implement novel and special convolutional methods on HSI data
%     \item Review the usage of newer technologies and architectures such as transformers and U-nets conformed to the Hyperspectral plane
%     \item Evaluate the use of Graph Based and Attention based networks and their performances in recent works
% \end{itemize}

\section{Datasets}
\label{section2}
The HSI datasets used by most methods involve land cover labels. The following datasets were the main testbeds used by the various models we look at in our survey. These datasets are predominantly shot via airborne sensors, and have applications in remote sensing. We look at each of these datasets in brief: 

\begin{enumerate}
    \item Indian Pines : The Indian Pines dataset \cite{PURR1947} was captured by an Airborne Visible Infrared Imaging Sensor (AVIRIS) equipped on a plane over a test site in North West Indiana USA, covering a variety of crops, as well as some housing structures and highway roads. Consists of 145*145 pixels over 224 bands of which 200 are kept after removing bands covering water absorption. A total of 16 class labels are available.

    \item Salinas : The Salinas dataset was captured over the Salinas Valley in California USA by an AVIRIS Sensor equipped on a plane by NASA on AVIRIS\cite{grupo-de-inteligencia-computacional-2021}. The image of 512*218 pixels, has 200 bands out of 224 after water absorption bands are discarded. Salinas contains 16 ground-truth classes, which usually consist of type of vegetation or plants, such as fallow, stubble, brocoli and so on. 

    \item WHU-Hi-HongHu : The WHU-Hi-HongHu dataset \cite{hu2021whuhi} is an airborne shot of the Honghu City, Hubei Province, China, shot with a 17-mm focal length Headwall Nano-Hyperspec
imaging sensor equipped on a DJI Matrice 600 Pro UAVplatform. The imagery of size 940*475 pixels provides 270 bands from the range 400nm to 1000nm. There are 11 ground-truth labels consisting of type of crops coupled with non-crop material types, such as Cotton, Cabbage and so on. 

    \item Pavia University : The Pavia University Dataset\cite{telecommunications-and-remote-sensing-laboratory-pavia-university-no-date} is a 610*610 pixel image captured with the Reflective Optics Imaging Spectrometer (ROSIS) shot by a plane over the city of Pavia Italy. The sensor captures 103 bands with 9 ground-truth classes consisting of material types such as Asphalt or Trees and so on.
    
    \item Houston University : The Houston University Dataset is a 1905*349 pixel image captured with a spatial resolution of 2.5m per pixel through the ITRES-CASI 1500 sensor for the purposes of the 2013 IEEE GRSS Data Fusion Contest\cite{iee-grss-data-fusion-2013}. The sensor captures 144 bands and yields 15 ground-truth classes over the landscape of the University of Houston and the neighbouring urban area.
\end{enumerate}


\begin{table*}[t]
\small
\caption{Dataset Statistics}
\label{tab:data1}
\resizebox{\textwidth}{!}{%
\begin{tabular}{lrrlllrrr}
\hline
\multirow{2}{*}{Datasets}                                     & \multicolumn{1}{l}{\multirow{2}{*}{Classes}} & \multicolumn{1}{l}{\multirow{2}{*}{Bands}} & \multirow{2}{*}{Dimension} & \multirow{2}{*}{Sensor}                                                & \multirow{2}{*}{Type} & \multicolumn{1}{l}{\multirow{2}{*}{\begin{tabular}[c]{@{}l@{}}Average Train\\ Pixels\end{tabular}}} & \multicolumn{1}{l}{\multirow{2}{*}{\begin{tabular}[c]{@{}l@{}}Average Test\\ Pixels\end{tabular}}} & \multicolumn{1}{l}{\multirow{2}{*}{\begin{tabular}[c]{@{}l@{}}Total Labeled\\  Pixels\end{tabular}}} \\
                                                              & \multicolumn{1}{l}{}                         & \multicolumn{1}{l}{}                       &                            &                                                                        &                       & \multicolumn{1}{l}{}                                                                                & \multicolumn{1}{l}{}                                                                               & \multicolumn{1}{l}{}                                                                                 \\ \hline
Pavia University   & 9                                            & 103                                        & 610*610                    & ROSIS                                                                  & Non Crop              & 734.8                                                                                               & 42041.2                                                                                            & 42776                                                                                                \\
Indian Pines       & 16                                           & 220                                        & 145*145                    & AVIRIS                                                                 & Crop + Misc           & 581                                                                                                 & 9565.5                                                                                             & 10249                                                                                                \\
Salinas Scene                                                       & 16                                           & 220                                        & 512*218                    & AVIRIS                                                                 & Crop + Age            & 1059.34                                                                                             & 53057.67                                                                                           & 54129                                                                                                \\
WHU-Hi-HongHu                                                 & 22                                           & 270                                        & 940*475                    & \begin{tabular}[c]{@{}l@{}}Nano-Hyperspec \\ Image Sensor\end{tabular} & Crop                  & 784.5                                                                                               & 184142                                                                                             & 196271                                                                                               \\
Houston University & 15                                           & 144                                        & 1905*349                   & CASI                                                                   & Non Crop              & 1904                                                                                                & 13125                                                                                              & 15029                                                                                                \\ \hline
\end{tabular}%
}
\end{table*}

\begin{table}[t]
\caption{Dataset usage in assessed models}
\label{tab:dat2}
\resizebox{\columnwidth}{!}{%
\begin{tabular}{lllllll}
\toprule
\multirow{2}{*}{Datasets}                                     & \multicolumn{6}{c}{Usage in Models}              \\ \cmidrule(l){2-7} 
                                                              & RIAN & SSFTT & MFGCN & EMS-GCN & CWG-SAGE & WFCG \\ \midrule
Pavia University   &  \Checkmark & \Checkmark & \Checkmark  &  \Checkmark &      &  \Checkmark \\
Indian Pines   &    & \Checkmark  &  & \Checkmark &    \Checkmark   &  \Checkmark  \\
Salinas Scene &  \Checkmark &   &  \Checkmark & &  \Checkmark &  \\
WHU-Hi-HongHu                                                 &                           &                            &                            &                              &              \Checkmark                  &        \Checkmark                    \\
Houston University &      \Checkmark                      &    \Checkmark                         &   \Checkmark                          &    \Checkmark                           &                               &                           \\ \bottomrule
\end{tabular}%
}
\end{table}

\section{Deep Learning Based Methods}
\label{section3}

Machine Learning methods such as Support Vector Machines have been used dominantly in the past for HSI classification tasks. However they suffer from lack of generalizability and may often not achieve state of the art results\cite{zhang2022ems} \cite{paoletti-2019}. In comparision, Deep Learning methods have shown their ability to achieve much superior results and in general can utilize the techniques used in current vision intelligence to conform to the hyperspectral plane. 

Graph Convolutional Network(GCN)\cite{DBLP:journals/corr/KipfW16} based methods\cite{cao-2022} and Attention based methods\cite{guo-2022} utilize newer techniques that show great promise for computer vision. GCN's first emerged as a semi-supervised technique to adapt convolutional techniques on a graph node based on the features of it's surrounding connected nodes in conjunction with it's own features. As opposed to adjacency matrix based convolution methods, GCN's dealt with the sparsity of the adjacency matrices and came up with a way to preserve a node's independent features while also taking into account the features of it's neighbouring nodes through the use-case of node embeddings. Many questions arise when it comes to the application of GCN's into vision based methods. As a method that mainly conforms to non-Euclidean spatial data (graphs), how best can we translate Euclidean data such as images into non-Euclidean space for use with the GCN topography. 

Specific to Hyperspectral Images, the use of GCN's at a glance seems appealing mainly due to the dimensionality of the 100+ bands usually associated with a Hyperspectral image. There are similarities between the band makeup of a single pixel consisting of data across a spectrum of 100+ wavelengths, and node features that we commonly associate with a graph node. It offers a way to sidestep the dimensionality problem faced by simple vision convolutional operators that require dimensionality reduction or the use of expensive 3D convolutions. A common use case observed across the papers surveyed to tackle the conversion of the Euclidean HSI cube to non-Euclidean graph nodes is to segment the cube into superpixels that act as supergraphs\cite{cheng-2022}. Most of the Graph based methods observed implement a fusion of graph superpixel based features with pixel-level base features for classification. 

Attention\cite{vaswani-2017}  as a technique has been revolutionary in enabling learning models to learn much more than what was thought possible. The attention mechanism was originally\cite{bahdanau-2014} used to tackle the bottleneck of conventional encoder-decoder models by assigning associated energy to hidden variable length annotation vectors. The approach of assigning weights to the weights of an input to compute the importance of certain segments was one that was quickly adapted into a number of different use-cases. In terms of applications in vision, we have multiple forms of attention that plays a part in improving classic vision transformer models, as discussed in detail in Section \ref{att}.

\begin{figure*}[hbt!]
    \centering
    \includegraphics[width=0.8\textwidth]{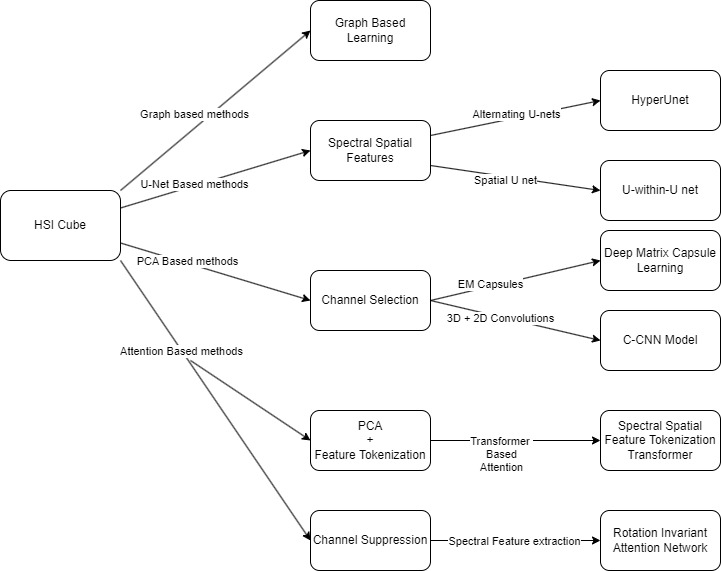}
    \caption{Overview of model methodologies}
    \label{fig:overview}
\end{figure*}

In the following sections, we discuss the methodologies of models with respect to the various modules, as per Figure \ref{fig:overview}.

% \subsection{Special Convolutional Techniques}

% While the popular method of convolution over images is widely used and utilized in vision, the intermixing of spatial and spectral features in convolutional networks can lead to inaccurate results. To combat this drawback and attune convolutional techniques to HSI, many different methods of convolutions have been applied to HSI Learning. 

% \subsubsection{Deep Matrix Capsules}

% The usage of capsule learning \cite{sabour-2017} is based on the concept of using a pose value as a vector rather than a scalar value to capture the orientation of the features with respect to each other.  

% The capsules utilize EM routing to finalize the pose parameters and capsule activations in a capsule convolutional layer. Uses two alternating steps called E-step and M-step to fit a mixture of Gaussian distributions to the data.

% \subsubsection{Consolidated Convolutional Neural Network}

% \begin{figure}
%     \centering
%     \includegraphics[width=0.4\textwidth]{figures/ccnn.jpg}
%     \caption{C-CNN architecture}
%     \label{fig:my_label}
% \end{figure}
%  The Consolidated Convolutional Neural Network (C-CNN) \cite{sun-2022} uses a 3D convolution along with a 2D convolution to extract spectral and spatial information without intermixing of the features. Additionally the use of one 3D and one 2D convolution leads to a much quicker compute time. 

\subsection{Attention Based Methods}

\label{att}
The concept of Attention \cite{vaswani-2017} was a revolutionary advent in the field of Natural Language, enabling correlation and encoder decoder learning over much longer input output sizes. Both Encoder Decoder attention and Self Attention have been effective in leveraging correlation between hidden states of the input sequence, and the concept of self attention has also been explored in application towards assigning weights to the informative bands in the spectral makeup \cite{Mou-2020} \cite{Zheng-2022} \cite{fang-2019} and for spectral suppression.

Attention can also be used in the transformers used to encode and decode HSI images in various forms. We look at some recent works that leverage attention modules in tandem with recent techniques to produce better predictions.

The main forms of attention that concern HSI specifically, are Channel Attention and Spatial Attention(Position). Channel Attention refers to assigning different weights to the multiple channels of an image before it undergoes convolution. This is even more so relevant in HSI cubes wherein Principal Component Analysis(PCA) is often applied to reduce the dimensionality. With the help of Channel Attention the model suppresses certain bands that are less informative while enhancing the relevant ones. Through backpropagation it is ensured that the channel weights are relevant to the use-case of classification. Spatial or Position Attention refers to assigning different weights to the surrounding pixels so as to enhance some and suppress others during convolution. Used in tandem with Channel Attention, Spatial Attention displays the potential to attune better to the HSI input\cite{dong-2022}. 

\subsubsection{Rotation Invariant Attention Network}
\begin{figure*}[!hbt]
    \centering
    \includegraphics[width=0.8\textwidth]{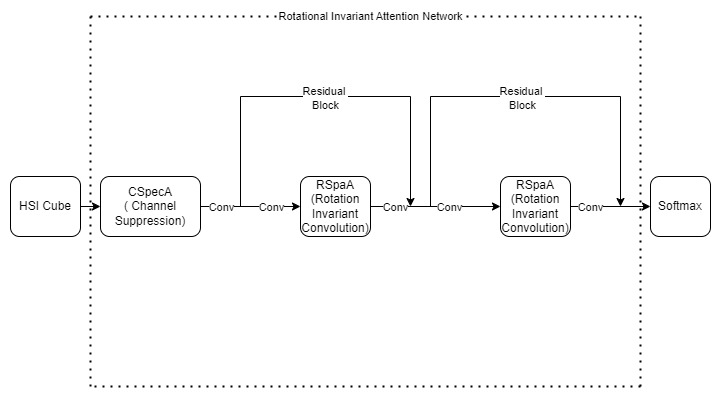}
    \caption{RIAN Model architecture}
    \label{fig:RIAN}
\end{figure*}

\hfill\\
The Rotation Invariant Attention Network (RIAN) \cite{Zheng-2022} aims to combine the use of an Attention Module with the use of a Rotation Invariant Convolutional network. The Attention module generates weights for the bands to recalibrate the spectral bands of individual HSI patches. The module also avoids the influence of interfering pixels around the center pixel that needs to be labelled with the help of generation of attention weights using the center pixel. This is combined with the Rotation Invariant Spatial Feature module to adaptively aggregate pixel features into rotation invariant spectral spatial features for convolution. The model, with all it's modules, are shown in Figure \ref{fig:RIAN}.

Another innovation lies in the Rotation Spectral Aggregation module (RSpaA), wherein instead of spectral spatial convolution a rotation invariant spectral feature extractor is followed by a spatial aggregation. A spectral convolution over the patch yields spectral features. During aggregation, only spectral features of pixels with high similarity are aggregated, with the help of a similarity weight based on a similarity threshold. 

With the help of the attention modules there is no data loss of spectral channel, as in comparision to other methods that utilize PCA or LDA. Furthermore, with the help of rotation invariant networks the information of adjacent pixels is properly assessed as in the case of land cover problem statements, translating exceptionally well to accuracy.

The use of an RSpaA module in aggregation of neighborhood features resembles in intuition a GCN. The rectified use of a GCN in tandem with the RSpaA module could result in better results and avoidance of effect of interfering pixels.

\subsubsection{Spectral Spatial Feature Tokenization Transformer}

In the Spectral Spatial Feature Tokenization Transformer (SSFTT) the HSI cube patches are converted into a flattened sequence of tokens with the help of a Spectral Spatial Feature extractor, enabled with a 3D convolution followed by a 2D convolution not unlike the C-CNN model \cite{sun-2022} to extract spectral spatial token features. 

\begin{figure*}[!hbt]
    \centering
    \includegraphics[width=0.8\textwidth]{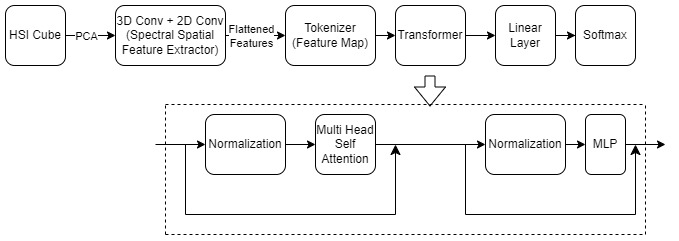}
    \caption{SSFTT Model architecture}
    \label{fig:SSFTT}
\end{figure*}

With the concatenation of a learn-able token that corresponds to the class label, the transformer is trained to regenerate the feature sequence with the class label at the front. The transformer uses Multi-Head and regular Self Attention \cite{vaswani-2017} to make the model attune to all bands and the long feature size of the sequence. At evaluation time a linear softmax classifier coupled with the final vector is used to obtain the label, as shown in Figure \ref{fig:SSFTT}.

Attention based methods use a feature token space for both spectral and spatial features of 
 a Hyperspectral Image. This helps the model avoid excessive spectral spatial feature mixing and lets the model attune to different token spaces for spectral and spatial features, rather than one set of contiguous activation values.

\subsection{U-net based methods}
\begin{figure}
    \centering
    \includegraphics[width=0.4\textwidth]{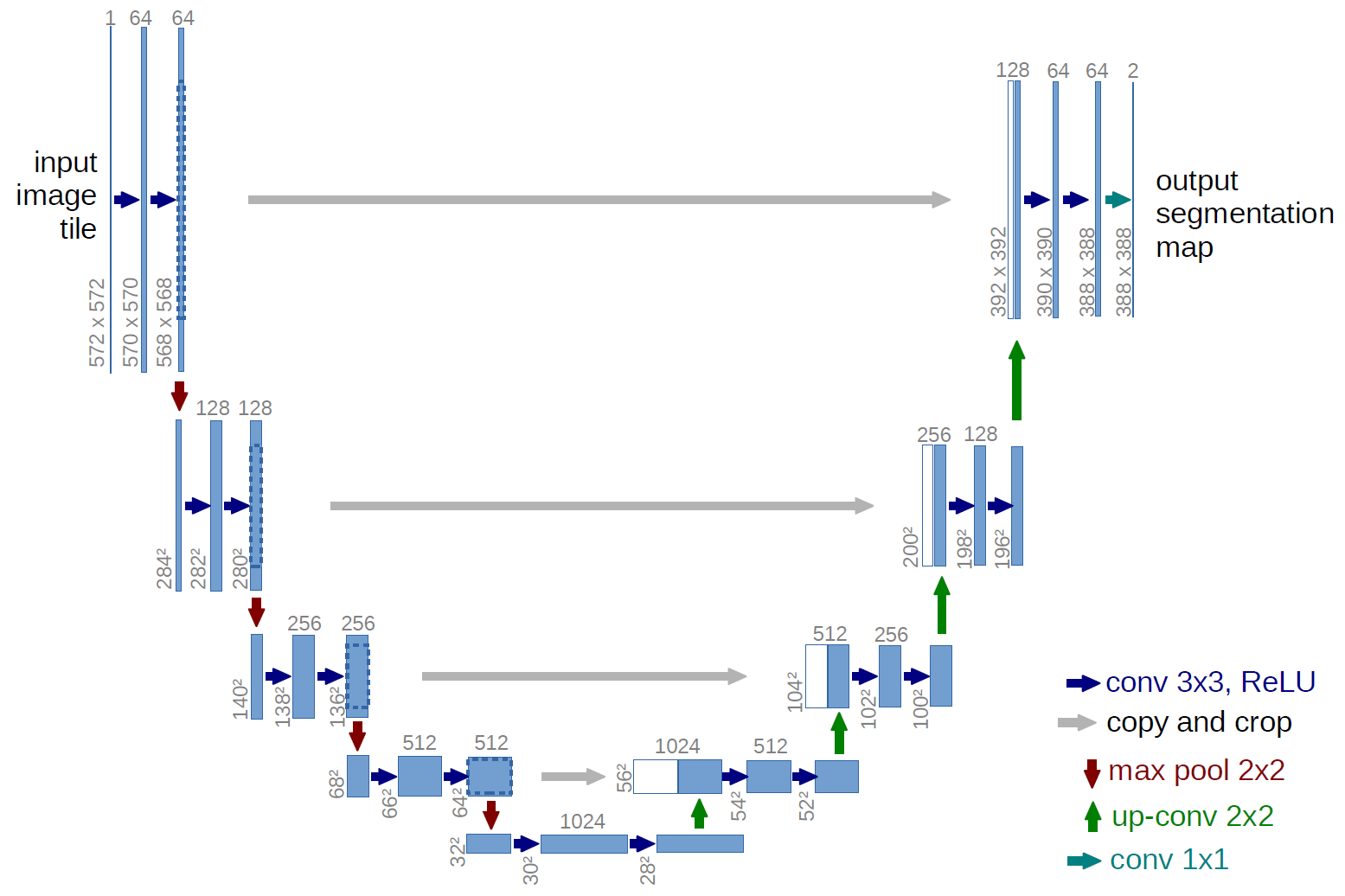}
    \caption{U-net architecture}
    \label{fig:UNet}
\end{figure}

U-net \cite{RFB15a} functions as a specialized auto-encoder that encodes and decodes  spatial information in an input image, to yield 
a new output image. This is useful in cases wherein we have an a mask of output class as in pixelized Image Classification or segmentation. U-net is different from a traditional auto-encoder in that it utilizes skip connections for recontextualization of information at equal  encode-decode levels, as shown in Figure \ref{fig:UNet}. 

However, as U-nets are designed for conventional images, there are many ways of conforming them to the Hyperspectral domain. 

\subsubsection{HyperUNet}

\hfill\\
 In the HyperUNet \cite{paul-2021} the contiguous spectral bands are preserved by partitioning the Hyperspectral cube into two partitions with the selection of alternate bands. Each partition is processed with identical U-net architectures, with the final outputs concatenated to yield the segmented image. The paper yields significant improvements over the use of a traditional U-net, and offers a different outlook on spectral segmentation, with the use of a Depthwise Convolution rather than a traditional convolution to reduce the number of parameters and reduce computational complexity. 

 However, the use of spectral partitioning to yield two different outputs that are finally concatenated might still result in the intermixing of spectral and spatial information, resulting in inaccurate results.

\subsubsection{U-within-U Net}

\hfill\\
 The U-within-U net \cite{manifold-2021} aims to provide a similar outlook of using two U-nets, however they take a different approach to the problem of spectral and spatial learning. 

\begin{figure}[!hbt]
    \centering
    \includegraphics[width=1\columnwidth]{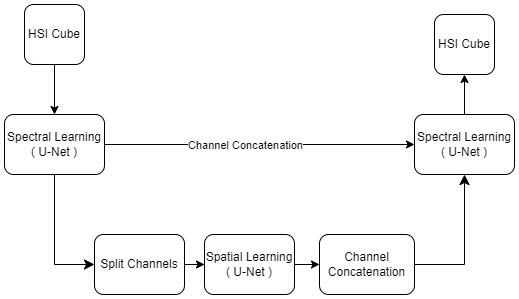}
    \caption{U-within-U Net}
    \label{fig:uWu}
\end{figure}

The U-within-U net (UwU-Net) aims to assign multiple separate U-nets, one specifically for the task of spatial learning and a collection of U-nets specifically for the task of spectral learning, as shown in Figure \ref{fig:uWu}. This architecture dedicates tunable free parameters to both spectral information and spatial information, resulting in a holistic approach that tackles the issue of spectral and spatial intermixing effectively.

\subsection{Graph Convolutional Network Based Methods}

Graph Learning is a booming field with the advent of Big Data, and the introduction of Graph Convolutional Networks \cite{DBLP:journals/corr/KipfW16} has brought improvements in strides for Graph Deep Learning. The intuition behind GCN, that learns the class or output label of a node only based on the node's own features and the features of it's neighbours in a convolutional manner, can also be applied to HSI wherein the pixel can be viewed as a node with the spectral values as it's feature make-up.

\begin{figure*}[!hbt]
    \centering
    \includegraphics[width=1\textwidth]{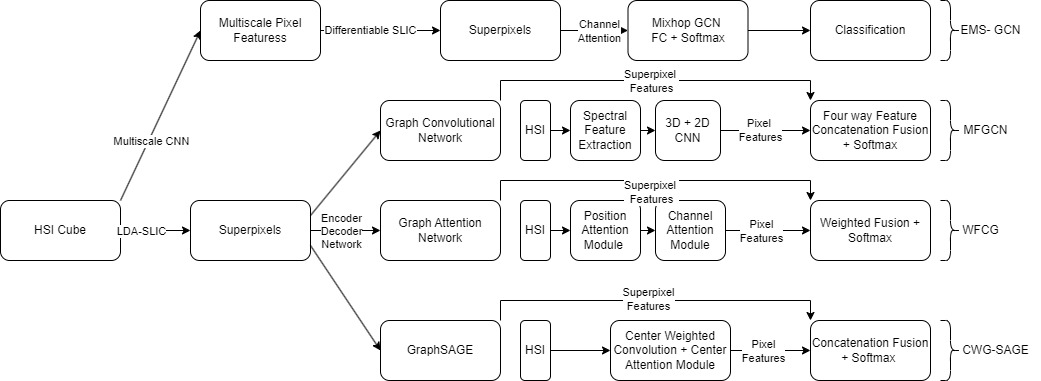}
    \caption{Graph Based Method Modular Overview}
    \label{fig:GCN}
\end{figure*}

Using a semi-supervised method of learning to attune the entire HSI cube to a few key pixels or regions is the main intuition that helps the GCN model conform well to HSI and helps use the spectral information as feature vectors that provide vital and key information. This section attempts to analyze and contrast the approaches taken by various Graph Learning based methods, including a modular make-up as shown in Figure ~\ref{fig:GCN}.

\subsubsection{Multi Feature Fusion GCN}

\hfill\\

The Multi Feature Fusion Graph Convolutional Network (MFGCN) \cite{ding-2022} aims to adapt both classic convolutional learning and graph convolutional learning to effectively classify HSI pixels. 

The HSI cube is partitioned into adaptive regions termed super-pixels by means of Linear Discriminant Analysis(LDA) to reduce dimensionality followed by Simple Linear Iterative Clustering.
The node features of these super-pixels are constructed using 1D Convolutions over the spectral features of the super-pixel and it's member pixels. Thus pixels in a particular region are assigned their respective super-pixel feature vectors. 

Apart from GCN-based node features, each pixel is put through a 1D and 2D convolutional filter to obtain pixel specific features, which can then be concatenated with the super-pixel feature vector to yield a complete feature makeup. A softmax classifier applied over the final concatenated feature vector enables loss calculation and training to attune the fusion network over the HSI data. 

The use of a fusion of graph based convolutions to attune pixels to regions, and classical convolutions to gather spectral and spatial information offers a different outlook on feature extraction wherein rather than focus on spectral and spatial features being separate, the focus is on pixel to region and region to pixel assignment. However, the use of LDA to reduce dimensionality may cause data loss and a better method of dimensionality reduction could be explored. Furthermore, exploration of techniques other than concatenation of features in fusion could lead to better fusion. 

\subsubsection{End to End Mixhop GCN}
 The End to End Mixhop Superpixel based GCN \cite{zhang2022ems}  (EMSGCN) aims to incorporate novel methods in superpixel association and a mixhop GCN that focuses on zero-hop and one-hop neighborhood information simultaneously. 

 The Differentiable Superpixel Segmentation module implements a multiscale CNN module along with an SLIC module to gain super-pixels. This differs from simple SLIC application featured in MFGCN \cite{ding-2022}, WFGC \cite{dong-2022} and CWG-SAGE \cite{cui-2023}

\subsubsection{Weighted Feature Fusion Convolution and Graph Network}

Aiming to integrate the use of CNN for Euclidean based Deep Learning and GCN for non-Euclidean connections between different land covers, the Weighted Feature Fusion and Graph Attention Network \cite{dong-2022} (WFGC) proposes a novel framework for weighted fusion of a CNN model and a Graph Attention Network Model \cite{velickovic-2017}. The model applies attention over both the spectral and spatial domain to ensure a holistic combination of attention based and graph based methods.

The model applies two Attention mechanisms to introduce better correlation between the pixels and channels. The Position Attention Module acts as a spatial attention module to refine pixel-pixel relation, while the Channel Attention Module acts as a spectral attention module to refine the weights of channel feature maps.

SLIC is applied again to differentiate the image into super-pixels. Akin to a GAT, the use of an Encoder - Decoder mechanism is applied to convert grid features into node features and back. 

Rather than simple concatenation fusion, the WFGC employs a weighted sum of features to yield the final feature vector for prediction. 

\subsubsection{Center Weighted Convolution }
\hfill\\

The Center Weighted Convolution and GraphSAGE Network (CWG-SAGE) works in similar fashion to the MFGCN, with pixel-region assignment, pixel and feature level feature extraction and fusion to yield classification. The innovation introduced in their work lies in the convolutional module, where instead of a simple convolution, the center pixel gets higher weight so that the relationship between edge features and central features is reconsidered. 

The two key parts to this are the Centre Weighted module that uses an Asymmetric Convolutional Block \cite{DBLP:journals/corr/abs-1908-03930} coupled with the Centre Attention Module that uses the 3X3 block from the middle of the HSI block as the kernel to better gauge the relationship between centre and edge pixel features. 

The fusion ensures that the Pixel-Region assignment features are coupled well with the CW convolution module that captures the relationship between center pixels and it's edge pixel features well. As a result, the model performs better in terms of fusion learning. 

\begin{table*}[hbt!]
\caption{Detailed Benchmarking of SOTA GCN and Attention Networks for HSI Classification}
\label{tab:metrics}
\resizebox{\textwidth}{!}{%
\begin{tabular}{llcccccc}
\hline
\multirow{2}{*}{Datasets}         & \multirow{2}{*}{Metrics} & \multicolumn{6}{c}{Models}                                                                                                                                               \\ \cline{3-8} 
                                  &                          & \multicolumn{1}{l}{RIAN} & \multicolumn{1}{l}{SSFTT} & \multicolumn{1}{l}{MFGCN} & \multicolumn{1}{l}{EMS-GCN} & \multicolumn{1}{l}{CWG-SAGE} & \multicolumn{1}{l}{WFCG} \\ \hline
\multirow{3}{*}{Indian Pines}     & OA                       & -                        & 97.47                     & -                         & 95.87                       & \textbf{98.29}               & 90.86                    \\
                                  & AA                       & -                        & 96.57                     & -                         & \textbf{97.45}              & 96.09                        & 87.88                    \\
                                  & Kappa                    & -                        & 97.11                     & -                         & 95.27                       & \textbf{98.05}               & 89.58                    \\ \hline
\multirow{3}{*}{Pavia University} & OA                       & 98.66                    & 99.21                     & \textbf{99.49}            & 98.47                       & -                            & 93.47                    \\
                                  & AA                       & \textbf{99.32}           & 98.69                     & 99.2                      & 99.11                       & -                            & 91.65                    \\
                                  & Kappa                    & 98.48                    & 99.15                     & \textbf{99.32}            & 97.98                       & -                            & 91.37                    \\ \hline
\multirow{3}{*}{Salinas}          & OA                       & 97.75                    & -                         & 99.28                     & -                           & \textbf{99.63}               & -                        \\
                                  & AA                       & 99.13                    & -                         & 99.5                      & -                           & \textbf{99.69}               & -                        \\
                                  & Kappa                    & 97.49                    & -                         & 99.2                      & -                           & \textbf{99.59}               & -                        \\ \hline
\multirow{3}{*}{Houston}          & OA                       & 86.37           & \textbf{98.92}            & 95.24                     & 88.57                       & -                            & -                        \\
                                  & AA                       &88.68           & \textbf{99.01}            & 96.21                     & 89.92                       & -                            & -                        \\
                                  & Kappa                    &85.14          & \textbf{98.83}            & 94.85                     & 87.62                       & -                            & -                        \\ \hline
\multirow{3}{*}{WHU-Hi-HongHu}    & OA                       & -                        & -                         & -                         & -                           & \textbf{98.08}               & 93.08                    \\
                                  & AA                       & -                        & -                         & -                         & -                           & \textbf{96.94}               & 90.26                    \\
                                  & Kappa                    & -                        & -                         & -                         & -                           & \textbf{97.87}               & 92.25                    \\ \hline
\end{tabular}%
}
\end{table*}

\begin{table}[]
\caption{Training Dataset Size across multiple models}
\label{tab:datsize}
\resizebox{\columnwidth}{!}{%
\begin{tabular}{@{}lcccccc@{}}
\toprule
\multirow{2}{*}{Datasets} & \multicolumn{6}{c}{Train Dataset Size(\%) per Model} \\ \cmidrule(l){2-7} 
                          & RIAN   & SSFTT  & MFGCN  & EMS-GCN & CWG-SAGE & WFCG \\ \midrule
Indian Pines              & -      & 9.99   & -      & 4.39    & 6.15     & 2.15 \\
Pavia                     & 2.10   & 5.00   & 0.63   & 0.63    & -        & 0.22 \\
Salinas                   & 2.96   & -      & 0.89   & -       & 2.03     & -    \\
Houston                   & 18.84  & 9.99   & 2.99   & 18.84   & -        & -    \\
WHU-Hi-HongHu             & -      & -      & -      & -       & 0.69     & 0.21 \\ \bottomrule
\end{tabular}%
}
\end{table}

\section{Results and Discussion}
\label{section4}
% \usepackage{booktabs}
% \usepackage{multirow}
% \usepackage{graphicx}

All the models have been trained on and characterized on various datasets explored. Metrics include the Overall Accuracy (OA), the Average Accuracy (AA) and the Kappa Coefficient. In Table 3 \ref{tab:metrics}, we present additonal benchmarking of all models. 

\subsection{Performance Metrics}

Graph Based approaches have better performance in terms of OA, AA and Kappa in general. CWG-SAGE, using the GraphSAGE variant of GCN's, performs best on Indian Pines, Salinas and WHU-Hi-HongHu datasets. Most of the fusion-graph based methods, such as MFGCN, and CWG-SAGE perform very well, posting best or near-best performances across the table. While the SSFTT tokenizer transformer based model works well for all datasets and posts the best performance on the Houston dataset, The RIAN model does not maintain good performance for all datasets, failing to do as well as other models on the Houston dataset. 

\begin{figure}[hbt!]
    \centering
    \includegraphics[width=0.52\textwidth]{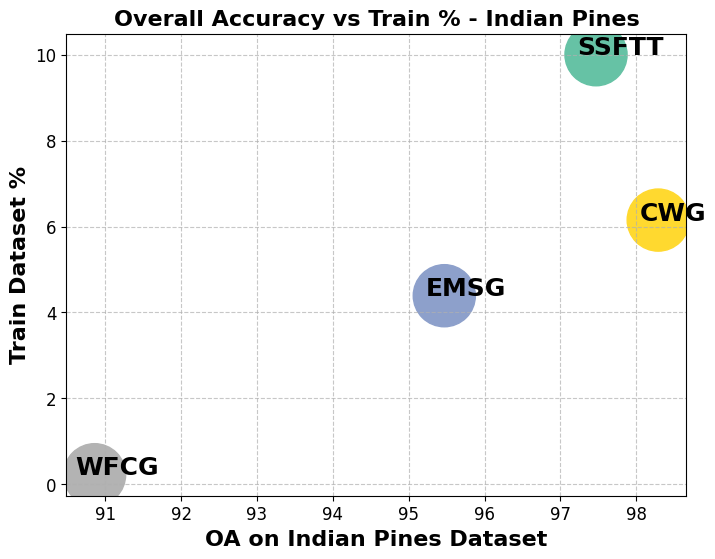}
    \caption{Train Data v OA Indian Pines}
    \label{fig:eval1}
\end{figure}

\begin{figure}[hbt!]
    \centering
    \includegraphics[width=0.5\textwidth]{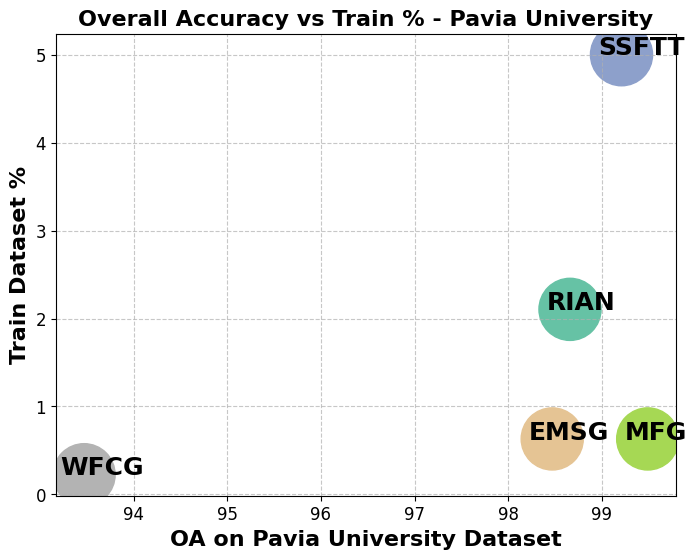}
    \caption{Train Data v OA Pavia University}
    \label{fig:eval2}
\end{figure}

\begin{figure}[hbt!]
    \centering
    \includegraphics[width=0.5\textwidth]{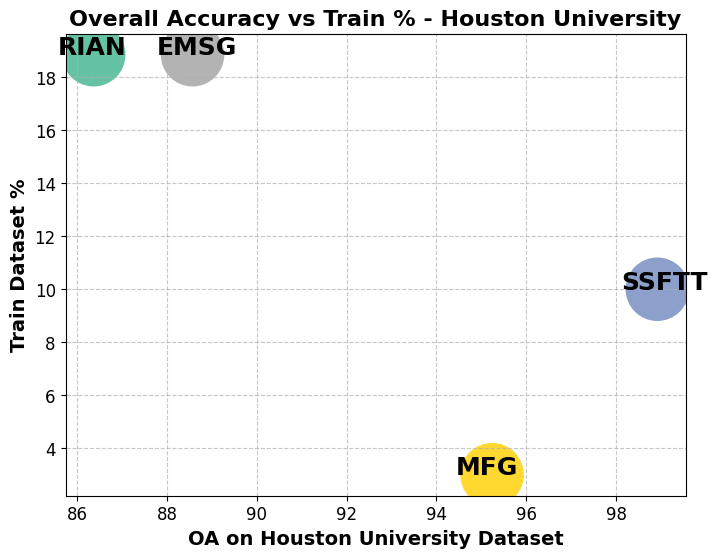}
    \caption{Train Data v OA Houston University}
    \label{fig:eval3}
\end{figure}

\begin{figure}[hbt!]
    \centering
    \includegraphics[width=0.5\textwidth]{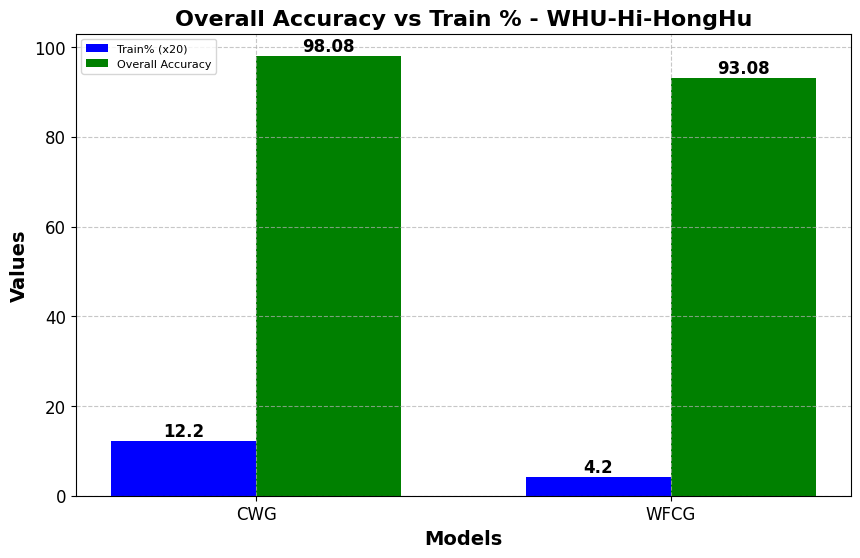}
    \caption{Train Data v OA Whu-Hi-HongHu}
    \label{fig:eval4}
\end{figure}

\begin{figure}[hbt!]
    \centering
    \includegraphics[width=0.5\textwidth]{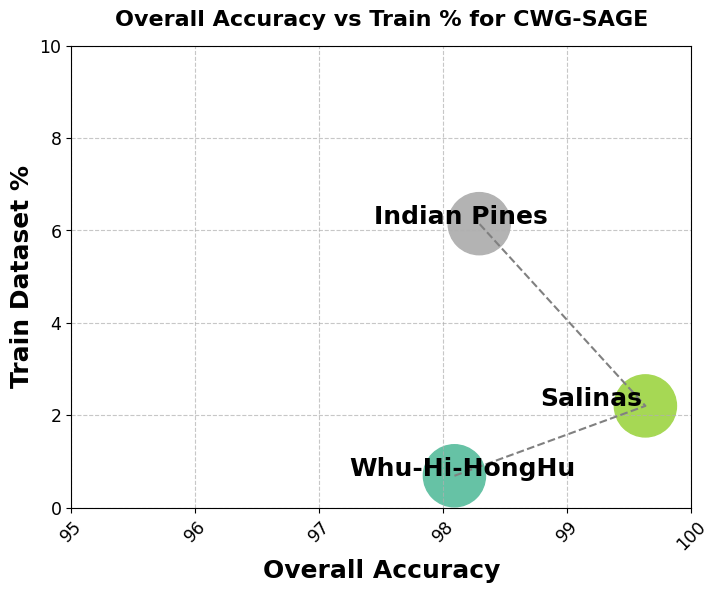}
    \caption{OA v Train Data for a GCN Based Method (CWG-SAGE)}
    \label{fig:oatgcn}
\end{figure}

\begin{figure}[hbt!]
    \centering
    \includegraphics[width=0.5\textwidth]{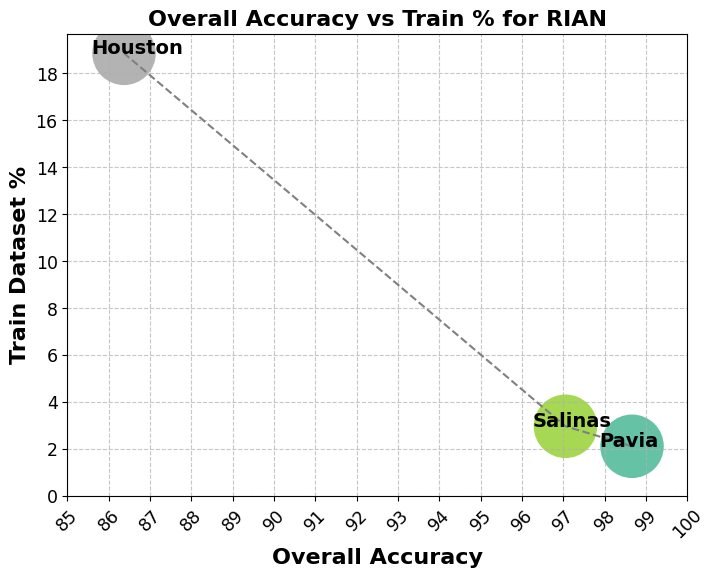}
    \caption{OA v Train Data for an Attention Based Method (RIAN)}
    \label{fig:oatatt}
\end{figure}

\subsection{Overall Accuracy with respect to Train Percentage on a common dataset}
We also look at the amount of data used as training data compared with the Overall Accuracy on multiple datasets, informed by the Overall Accuracy metrics described in Table \ref{tab:metrics} and the size of train dataset that the models trained on, described in Table \ref{tab:datsize}. Evaluating the graph on Indian Pines Dataset in Figure \ref{fig:eval1} .  While attention-transformer based methods have needed more data to train on, the GCN based methods have trained on only 1 percent of trainable pixel data. At the same time, GCN based methods show greater accuracy, showing a tendency of needing less data to give better results due to the fusion with global superpixel-based features in conjunction with local pixel-based features. CWG-SAGE shows the best performance while having used close to the least amount of train data pixels. We see that by supplementing pixel features with a fusion of super-pixel features, we can considerably reduce the amount of labelled pixel points needed to train a robust model.

The same plot can also be construed for the Pavia University Dataset as shown in Figure \ref{fig:eval2}. We see a trend wherein one method of Graph based fusion fares well(MFGCN) while the other fares poorly(WFCG). While usage of GCN and GraphSAGE fares well on both Indian Pines and Pavia University datasets (CWG-SAGE and MFGCN), the use of a Graph Attention Network in WFCG does not yield results of the same accuracy level. However, the performance of WFCG on incredibly low percentage of trainable data utilized is remarkable, hardly utilising more than 0.3 \% data for Pavia University, and more than 3 \% data for Indian Pines, while yielding close to best results for all. Further experimentation into the utilization of trainable data to yield better results is definitely something that can be considered given the performance on very little data. 

We compare methods that both utilize GCN in the Figure \ref{fig:eval4}, where the CWG-SAGE and WFCG models perform with a distinct difference in Overall Accuracy. While the difference in data used (0.69 : 0.21) favours CWG-SAGE, we see that WFCG frequently performs below SOTA. WFCG also uses much less data than any of it's counterparts as can be observed in Table \ref{tab:datsize}.

Finally, upon observation of the Houston University, Figure \ref{fig:eval3} yields an interesting phenomenon wherein the models that utilized less training data appeared to perform better on the test accuracy, with both SSFTT and MFGCN performing better than their counterparts RIAN and EMS-GCN, which supports the claim of SSFTT and MFGCN performing better overall compared to EMS-GCN and RIAN. We see the reappearance of the trend supporting fusion models like MFGCN for performing well while using less labelled pixels. To further support the claim of GCN-based methods working well on less labelled data, Subsection \ref{subsection4.3} looks at performance of the same model on different datasets and different training dataset sizes.

\subsection{Overall Accuracy with respect to Train Percentage on a common method}
\label{subsection4.3}
We construct graphs that look at the relationship between Overall Accuracy (OA) and the percentage of Train Dataset used for different datasets given a particular GCN-based method and Attention-based method. Upon observation, we see that while Attention Based methods suffer linearly over Train dataset percentage and consequent performance, GCN Based methods are able to perform for varying train dataset percentages used. As we observe, Figure \ref{fig:oatatt} yields a linear relation between the amount of dataset used and the Overall Accuracy, whereas in Figure \ref{fig:oatgcn} does not show the same correlation. We see from the graph clearly that GCN performance does not decrease with a decrease in training data. One possible conclusion to be surmised from this trend may be that the utilization of super-pixel features makes up for the lack of data, therefore bolstering GCN-based methods to situations in which labelled training data is scarce.

\section{Conclusion}

In this survey, we assessed various methods in which Graph-based and Attention-based models have been used to complement complex vision classification techniques associated with Hyperspectral Imaging.  We looked at five popular HSI datasets and highlighted differences in each via tabular form(See Table \ref{tab:data1}).We assessed the use of U-nets and their specialized autoencoder methods.

We observed multiple trends, including the ability of fusion GCN-based methods to adapt to data and achieve convergence with fewer labelled data points. We also highlighted the usage of Attention mechanisms and their effective utilization in achieving good results in comparision to GCN-based methods. This survey aimed to bring focus to a brewing trend of utilization of non-vision techniques in complex HSI vision classification utilization.

\pagebreak

\bibliographystyle{ACM-Reference-Format}
\bibliography{sample-base}

% \appendix









\end{document}